# CROWD MANAGEMENT, CRIME DETECTION, WORK MONITORING USING AI/ML

Manoj Kumar .R
*Assistant Professor, School of Computer Science and Engineering*
*Vellore Institute of Technology,*
Chennai, India
Manojkumar.r@vit.ac.in

P.R.Adithya
*UG Scholar, School of Computer Science and Engineering*
*Vellore Institute of Technology*
Chennai, India
Adithya.pr2022vitstudent.ac.in

Akash
*UG Scholar, School of Computer Science and Engineering*
*Vellore Institute of Technology*
Chennai, India
akash.b2022@vitstudent.ac.in

*Dheepak*
*UG Scholar, School of Computer Science and Engineering*
*Vellore Institute of Technology*
Chennai, India
dheepak.s2022@vitstudent.ac.in

HARSHINI .V
UG Scholar,School of computer science and engineering
vellore institute of technology
chennai,india
harshini.v2022@vitstudent.ac.in

SAI LAKSHANA
ug scholar,school of computer science and engineering
vellore institute of technology
chennai,india
sailakshana@vitstudent.ac.in

*Abstract:*

*This research endeavors to harness the potential of existing Closed-Circuit Television (CCTV) networks for a comprehensive approach to crowd management, crime prevention, and workplace monitoring through the integration of Artificial Intelligence (AI) and Machine Learning (ML) technologies. The primary objective is to develop and implement advanced algorithms capable of real-time analysis of video feeds, enabling the identification and assessment of crowd dynamics, early detection of potential criminal activities, and continuous monitoring of workplace environments. By leveraging AI/ML, the project aims to optimize surveillance capabilities, thereby enhancing public safety measures and improving organizational productivity. This initiative underscores the transformative impact that intelligent video analytics can have on existing infrastructure, mitigating the need for extensive system overhauls while significantly advancing security and operational efficiency.*

## I. Introduction

In our project, we leveraged the power of teachable machines to revolutionize crime detection and crowd management. Harnessing the potential of artificial intelligence, particularly in computer vision, we employed advanced algorithms to analyze image inputs, including CCTV videos, with unprecedented precision. These CCTV videos have been converted to images in .jpg format using an online tool, and these converted images were used to train the model. The model employs a Convolutional Neural Network (CNN) to classify the input video as either normal or indicative of criminal activity. Furthermore, our application extends beyond crime prevention to crowd management, where the same technology facilitates real-time monitoring and analysis of crowded spaces. The teachable machines not only enable us to identify potential security threats but also provide valuable insights for optimizing public safety strategies. This project represents a significant step forward in the fusion of cutting-edge technology and public safety, showcasing the potential of artificial intelligence in creating safer and more secure environments for communities.

**Now to know, how a convolution neural network works lets break it into parts. the 3 most important parts of this convolution neural networks are,**

1. **Convolution**
2. **Pooling**
3. **Flattening**

**Convolution**
Consider a 28x28 image, like those in the MNIST dataset used for handwritten digit recognition. In a basic artificial neural network setup, each pixel's value is treated as an individual feature input, resulting in 784 input nodes. While this approach may yield satisfactory results, it falls short in recognizing crucial features within the image. The model essentially processes each pixel independently, potentially missing important patterns.

Scaling this concept to a larger image, such as a 1920x1080 Ultra HD image, poses significant challenges. Applying the same methodology would result in an impractical 2 million input nodes. Even with a relatively modest hidden layer of 64 nodes, which is insufficient for such a large input, the network would involve a staggering 130 million weights. This massive scale of parameters necessitates an enormous computational load, overwhelming the capabilities of most machines. The sheer volume of calculations involved makes it unfeasible for effective image recognition, emphasizing the need for more sophisticated approaches in handling high-resolution images.

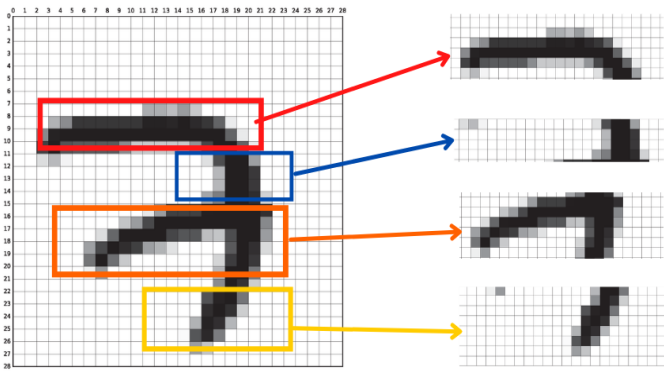

Important features of the image

A. To enhance image recognition in a neural network, it's crucial to identify and focus on the most relevant features while discarding unnecessary pixel information. Convolution, despite its reputation as a complex topic, is a fundamental technique that simplifies this process. In essence, convolution involves sliding a filter, or kernel, across an image to detect various features. Kernels, represented as 2D matrices with distinct weights, traverse the image, replacing pixel values with the weighted average of neighboring values.

B. This method allows us to pinpoint significant features in an image, providing a more nuanced understanding. By applying multiple randomly generated kernels, we can extract diverse features, enriching the model's ability to recognize intricate patterns.

C. Following the convolution layer, the next step involves pooling these features. Pooling is a technique used to condense and emphasize essential information while reducing computational complexity. This ensures that the model retains crucial details while discarding redundant information, contributing to improved image recognition performance.

D. *Pooling*
After identifying crucial features through convolution, the challenge remains in handling the large number of inputs, a task addressed by pooling. Pooling serves to shrink the image dimensions while retaining the features uncovered during convolution. Take, for instance, the MaxPooling method, which operates on a matrix shape and outputs the maximum value within that region. This technique allows for image compression without sacrificing the essential features, facilitating a more manageable input size for the subsequent layers of the neural network.

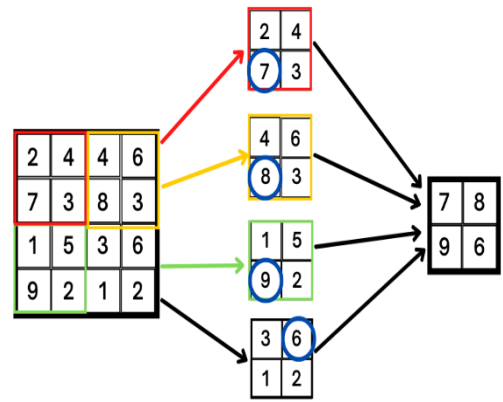

**MAX POOLING**

E. *Flattening*
Flattening is the process of transforming a 3D or 2D matrix into a 1D format, serving as the final step in preparing the image for input into the model. This step involves converting the structured representation of the image into a linear, one-dimensional input. The flattened data can then be seamlessly connected to a fully connected dense layer, facilitating subsequent stages of classification in the neural network

.

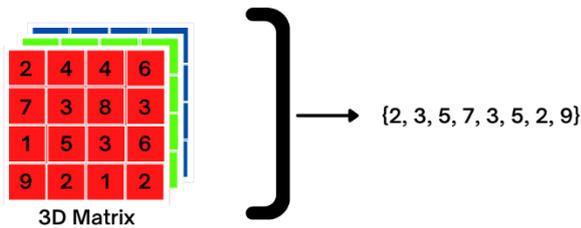

**Libraries used in this project:**
- OpenCV: Used to read the video input and splitting video into frames for analysing.
- Keras: Used to implement neural networks. It is a high-level neural network library that runs on top of tensorflow
- Numpy: Used to process images as the image pixel is in the form of matrix
- Pushbullet: It is an API used for sending SMS to mobile phone after detecting crime.

**Sample images used in Dataset:**

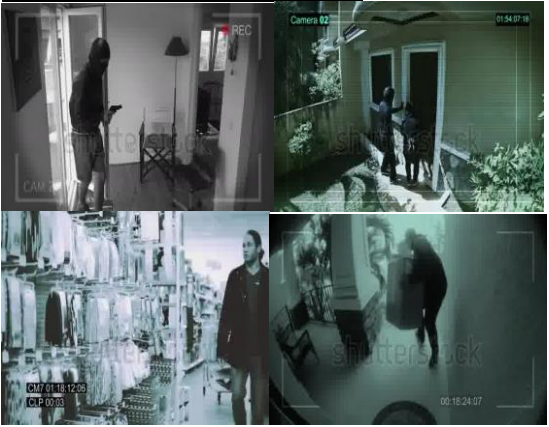

**Input &Output:**
**The input is a video that is being monitored for potential criminal activities. The output indicates whether the video involves suspicious activities or not. In the event of criminal behaviour, a notification is sent to the relevant authorities.**

## II. OBJECT DETECTION
A.EMPLOYEE MONITORING

In the dynamic landscape of contemporary workplaces, the need for efficient and effective work monitoring has become paramount. Organizations strive to optimize productivity, ensure employee safety, and maintain a secure work environment. The advent of machine learning technologies has opened up new avenues for addressing these challenges. This project report delves into the implementation of YOLO (You Only Look Once), a state-of-the-art object detection algorithm, as a pioneering solution for work monitoring.

### YOLO(YOU ONLY LOOK ONCE) MODULE

Unlike traditional object detection methods that involve multiple stages, YOLO streamlines the process, allowing for real-time detection with impressive speed and accuracy. Here's a step-by-step explanation of how the YOLO module works:

**Input Processing:** The input image undergoes grid-based division, forming the foundation for subsequent predictions.

**Bounding Box Prediction:** YOLO predicts multiple bounding boxes within each grid cell, each associated with parameters (x, y) for the box's center, width (w), height (h), confidence score, and class probabilities.

**Class Prediction:** YOLO determines the probability of each class for all bounding boxes in a grid cell, enabling simultaneous detection of multiple object classes in a given image.

**Confidence Score:** A confidence score indicates the model's certainty that a bounding box contains an object, with a range from 0 to 1.

**Non-Maximum Suppression:** Following predictions for all grid cells, a post-processing step, non-maximum suppression, removes redundant and low-confidence bounding boxes, retaining only the most confident and non-overlapping ones.

**Output:** The YOLO module produces a final output of bounding boxes, each linked to a class and a confidence score, representing the detected objects in the input image.

### INTERSECTION OVER UNION

Intersection over Union (IoU) is a metric used to evaluate the accuracy of an object detection algorithm, particularly in tasks such as image segmentation and bounding box prediction. IoU measures the overlap between the predicted bounding box and the ground truth bounding box for a given object in an image. The IoU is calculated as the ratio of the area of intersection between the predicted and ground truth bounding boxes to the area of their union. The formula for IoU is:

*IOU= Area of Intersection/Area of Union*

Here's a breakdown of the terms:

1. **Area of Intersection:** The region where the predicted bounding box and the ground truth bounding box overlap.

2. **Area of Union:** The combined region covered by both the predicted bounding box and the ground truth bounding box.

The IoU value ranges from 0 to 1, where:

IoU=0 indicates no overlap between the predicted and ground truth bounding boxes.

IoU=1 indicates a perfect overlap between the predicted and ground truth bounding boxes.

## III. LITERATURE SURVEY

1) *Uses OpenCV for object detection in computer Vision. LSTM (Long Short-Term Memory) is used to classify any event or behaviour as a crime or not. [Autonomous Anomaly Detection System for Crime Monitoring and Alert Generation]*

**Jyoti Kukad, Swapnil Soner, Sagar Pandya**

2) *Uses state-of-the-art face identification system Uses deepneural networks (DNN).*
*[Face Detection and Recognition for Criminal Identification System]*

**Sanika Tanmay, Aamani Tandasi, Shipra Saraswat**

3) *Uses Microsoft azure cognitive services and cloud system for implementation. Provides a comparative study of traditional methodologies used such as Haar Cascade.[ Proposed System for Criminal Detection and Recognition on CCTV Data Using Cloud and Machine Learning]*

**Samit Shirsat, Aakash Naik, Darshan Tamse**

4) *Uses pre trained deep leaning model VGGNet-19 which detects gun and knife. Uses SMS sending module to send alert.[ Crime Intention Detection System Using Deep Learning]*

**Umadevi V Navalgund, Priyadharshini.K**

5) *Focuses on identifying patterns and trends in crime occurrences. Uses ML and DL algorithms to predict crime related activities.[ Crime Prediction Using Machine Learning and Deep Learning: A Systematic Review and Future Directions]*

**VarunMandalapu , Lavanya Elluri**

## IV. Methodology

EMPLOYEE MONITORING

In the dynamic landscape of contemporary workplaces, the need for efficient and effective work monitoring has become paramount. Organizations strive to optimize productivity, ensure employee safety, and maintain a secure work environment. The advent of machine learning technologies has opened up new avenues for addressing these challenges. This project report delves into the implementation of YOLO (You Only Look Once), a state-of-the-art object detection algorithm, as a pioneering solution for work monitoring.

**YOLO(YOU ONLY LOOK ONCE) MODULE**

Unlike traditional object detection methods that involve multiple stages, YOLO streamlines the process, allowing for real-time detection with impressive speed and accuracy. Here's a step-by-step explanation of how the YOLO module works:

**Input Processing:** The input image undergoes grid-based division, forming the foundation for subsequent predictions.

**Bounding Box Prediction:** YOLO predicts multiple bounding boxes within each grid cell, each associated with parameters (x, y) for the box's center, width (w), height (h), confidence score, and class probabilities.

**Class Prediction:** YOLO determines the probability of each class for all bounding boxes in a grid cell, enabling simultaneous detection of multiple object classes in a given image.

**Confidence Score:** A confidence score indicates the model's certainty that a bounding box contains an object, with a range from 0 to 1.

**Non-Maximum Suppression:** Following predictions for all grid cells, a post-processing step, non-maximum suppression, removes redundant and low-confidence bounding boxes, retaining only the most confident and non-overlapping ones.

**Output:** The YOLO module produces a final output of bounding boxes, each linked to a class and a confidence score, representing the detected objects in the input image.

## V. RESULTS AND DISCUSSIONS

In the realm of advanced surveillance systems, the integration of work monitoring and crime detection has reached new heights, offering a comprehensive solution to enhance security measures. This innovative project leverages cutting-edge technologies, merging work monitoring outputs with crime detection capabilities, ultimately contributing to a safer and more efficient environment.

Upon capturing an input image indicative of theft or criminal activity, the system triggers an alert mechanism. This mechanism not only highlights the suspicious event but also sends an immediate alert message to designated authorities. The integration of heatmap visualization enhances the alert system by providing a visual representation of the anomaly, allowing authorities to swiftly assess the situation and respond effectively.

One of the project's standout features is the seamless integration of heatmap visualization. This graphical representation method offers a clear and intuitive display of numerical data, indicating the intensity of activities within the monitored space. In the context of work monitoring and crime detection, the heatmap becomes a powerful tool, showcasing the concentration and distribution of work hours and identifying anomalies that may indicate criminal behavior.

## VI. Conclusion

This project marks a significant advancement in the convergence of work monitoring and crime detection, offering a holistic solution that promotes both workplace efficiency and security. The synergy between advanced algorithms, specialized datasets, and heatmap visualization sets this system apart, exemplifying the potential of technology to revolutionize surveillance and safety measures in various domains. At the core of the system lies the utilization of sophisticated AI/ML algorithms, particularly the YOLO model, to simultaneously monitor work activities and detect criminal incidents. The YOLO model, renowned for its efficiency in object detection, ensures precise tracking of individuals and objects within the monitored space. The project's specialized dataset focuses on capturing both work-related scenarios and criminal activities, enabling the model to distinguish between routine work tasks and potential thefts.